\documentclass[letterpaper]{article} 
\usepackage{aaai2027}  
\usepackage[hyphens]{url}  
\usepackage{graphicx} 
\usepackage{natbib}  
\usepackage{caption} 
\usepackage{algorithm}
\usepackage{algorithmic}

\usepackage{newfloat}
\usepackage{listings}
\DeclareCaptionStyle{ruled}{labelfont=normalfont,labelsep=colon,strut=off} 
\floatstyle{ruled}
\newfloat{listing}{tb}{lst}{}
\floatname{listing}{Listing}

\usepackage{booktabs}

\usepackage{multirow}   
\usepackage{graphicx}
\usepackage{amsmath}
\usepackage{amssymb}
\usepackage{tcolorbox}

\lstdefinestyle{promptlisting}{
  basicstyle=\fontsize{6.0}{6.8}\selectfont\ttfamily,
  numbers=left,
  numberstyle=\tiny,
  numbersep=4pt,
  xleftmargin=1.5em,
  framexleftmargin=1em,
  frame=tb,
  rulecolor=\color{black},
  framerule=0.4pt,
  showstringspaces=false,
  tabsize=2,
  breaklines=true,
  breakatwhitespace=false,
  columns=fullflexible,
  keepspaces=true,
  captionpos=t,
  aboveskip=0.35\baselineskip,
  belowskip=0.35\baselineskip
}

\newcommand{\method}{ECAT}

\title{Entropy-based Code Adversarial Translation for Real-world Repository Migration}
\author{
    Yushun Tang\textsuperscript{\rm 1}, Yisen Cao\textsuperscript{\rm 1,2}, Zhicheng Chen\textsuperscript{\rm 1}, \\ Lin Peng\textsuperscript{\rm 1}, Junkang Mao\textsuperscript{\rm 1}, Fengyi Song\textsuperscript{\rm 1}, Yantao Jia\textsuperscript{\rm 1}\\
}
\affiliations{
    \textsuperscript{\rm 1}Huawei Technologies Co., Ltd.,
    \textsuperscript{\rm 2}The Chinese University of Hong Kong\\

    \tt\small \{tangyushun2, caoyisen, chenzhicheng17, \\ penglin44, maojunkang, songfengyi1, jiayantao\}@huawei.com
}

\begin{document}

\maketitle

\begin{abstract}
LLMs have demonstrated strong capabilities in code generation and automated program repair, but migrating an entire repository rarely produces a runnable application because long-horizon translation challenges LLM-based agents' ability to maintain repository-level migration objectives. In this work, we propose Entropy-based Code Adversarial Translation (ECAT), a multi-agent framework for automated Android-to-HarmonyOS repository migration. ECAT formulates repository migration as adversarial entropy minimization through a generator-discriminator architecture. The discriminator measures migration quality using a unified metric called Code Entropy and produces text gradients that specify both file-level generation directives and the skills needed to execute them. Guided by these optimization signals, the generator iteratively updates the repository, and each update is accepted only if it reduces Code Entropy. Repeated generator--discriminator interactions progressively drive the migration from an initial template toward a functionally complete HarmonyOS repository. Successful low-entropy trajectories are further distilled into a self-evolving memory tree, enabling transferable migration knowledge across repositories. We also introduce A2H-RepoBench, the first real-world benchmark for Android-to-HarmonyOS repository migration, covering applications from tens of thousands to hundreds of thousands of lines of code. Evaluated by node alignment and an agent-based functional judge, ECAT achieves 74.7\% overall migration quality and consistently outperforms existing agent-based methods across repositories of different scales.

\end{abstract}

\begin{links}
    \link{Code}{https://github.com/yushuntang/ECAT}
\end{links}

\section{Introduction}

Large language models (LLMs) have fundamentally advanced software engineering by enabling code generation and automated program repair from natural language and code repositories \cite{autocoderover2024,swebench2024,sweagent2024}. 
Among these tasks, repository migration remains particularly challenging because it requires long-horizon reasoning \cite{dong2026longhorizon} over large software repositories, where generation errors accumulate and propagate across long-range dependencies \cite{wang2025repotransbench}.

Android-to-HarmonyOS migration is an emerging representative repository migration task. Unlike file-level code translation or generation, it requires jointly transforming programming languages, UI frameworks, project configurations, and platform APIs while preserving repository-wide functionality. As repository size scales from tens to hundreds of thousands of lines of code (LOC), cross-file dependencies and long-range interactions rapidly increase, making globally consistent repository optimization increasingly difficult.
Despite recent progress, existing agentic software engineering systems still struggle at this scale. Long-context reasoning becomes less reliable as repository size grows, making it difficult to preserve cross-file consistency \cite{liu2024lost}. Most systems therefore rely on iterative self-refinement, where the same agent repeatedly generates and evaluates repository updates \cite{madaan2023selfrefine,shinn2023reflexion,yao2022react,openhands2024}. Such self-evaluation often overestimates intermediate repository quality \cite{zheng2023judging}, leading to premature convergence despite unresolved compilation errors, broken dependencies, or missing functionality. Moreover, migration experience from previous repositories is rarely accumulated, causing similar optimization trajectories to be repeatedly rediscovered from scratch.

Inspired by the adversarial learning paradigm of generative adversarial networks \cite{goodfellow2014gan}, in this work, we formulate repository migration as an adversarial entropy minimization problem. LLM-based code generation is inherently probabilistic, where migration errors and inconsistencies introduced during generation accumulate across files and modules, progressively increasing repository-level disorder. We characterize this disorder using \textit{Code Entropy}, where higher entropy indicates greater uncertainty, inconsistency, and unresolved migration errors within the generated repository. Following the principle of entropy minimization \cite{wang2021tent,agarwal2026unreasonable}, Code Entropy serves as a unified repository-level objective that measures repository correctness, migration fidelity, and runtime functionality. To optimize this objective, we propose Entropy-based Code Adversarial Translation (\textbf{\method{}}), a generator--discriminator framework that explicitly decouples repository generation from quality evaluation. The generator performs Android-to-HarmonyOS code generation, while an independent discriminator continuously evaluates repository quality, exposes high-entropy regions, and produces fine-grained text gradients \cite{yuksekgonul2024textgrad} that localize defects and recommend corresponding skills \cite{zheng2026skillrouter}. Guided by these optimization signals, the generator iteratively repairs the repository to reduce Code Entropy, forming an adversarial optimization loop in which the discriminator continually uncovers repository disorder while the generator progressively eliminates it. By separating generation from evaluation, \method{} enables sustained repository-level optimization instead of relying on the limited self-correction capability of a single agent. Successful low-entropy optimization trajectories are further distilled into a self-evolving memory tree, where migration knowledge is hierarchically organized, continuously refined, and transferred across repositories.

To facilitate systematic evaluation, we introduce A2H-RepoBench, the first real-world multiscale repository-level benchmark for Android-to-HarmonyOS migration. A2H-RepoBench comprises three representative real-world Android repositories spanning 50K, 120K, and 300K LOC, together with curated functional specifications and a unified evaluation protocol. These repository scales cover representative sizes of mature Android applications and introduce progressively more complex repository structures and cross-file dependencies, providing a realistic testbed for evaluating migration scalability. Experiments on A2H-RepoBench demonstrate that \method{} consistently outperforms representative agentic baselines in migration quality.
The major \textbf{contributions} of this work can be summarized as follows:
\begin{itemize}
    \item We formulate Android-to-HarmonyOS repository migration as an adversarial entropy minimization problem and introduce Code Entropy, a unified repository-level objective that quantifies repository disorder in terms of migration correctness and fidelity. Code Entropy provides an explicit optimization target for long-horizon repository migration.
    \item We propose \method{}, a generator–discriminator framework that explicitly decouples repository generation from repository evaluation. Instead of allowing an agent to judge its own outputs, an independent discriminator produces fine-grained text gradients that iteratively guide repository generation or repair. Successful trajectories are continuously distilled into a self-evolving memory tree, where migration knowledge is hierarchically organized and incrementally refined, enabling transferable experience across repositories.
    \item We construct A2H-RepoBench, the first repository-level benchmark for Android-to-HarmonyOS migration. A2H-RepoBench comprises three representative real-world Android repositories spanning 50K, 120K, and 300K LOC, together with functional specifications and a unified evaluation protocol. Extensive experiments on A2H-RepoBench demonstrate that \method{} consistently outperforms baselines in long-horizon migration quality.
\end{itemize}

\section{Related Work}

\paragraph{LLM-based agents.}

Large language models are increasingly deployed as autonomous agents that combine reasoning, tool use, memory retrieval, and environmental feedback. Early approaches relied on self-refinement, verbal reflection, and ReAct to improve a single agent's outputs \cite{madaan2023selfrefine,shinn2023reflexion,chen2023selfdebug,yao2022react}. More recently, multi-agent systems coordinate specialized agents for software development \cite{metagpt2023,chatdev2024}, test-driven programming \cite{agentcoder2023}, repository-level issue resolution \cite{sweagent2024,autocoderover2024,openhands2024,swebench2024}, and repair pipelines \cite{agentless2024}.
Recent surveys \cite{jiang2026selfimprovingagents} further identify self-improving agents as an emerging paradigm that continually evolves agent capabilities through accumulated experience during deployment. External memory has consequently emerged as a key mechanism for continual improvement without modifying model parameters. Existing memory-augmented agents store interaction histories, retrieved experiences, or reusable skills in external memories to improve future decision making
\cite{park2023generativeagents,packer2023memgpt,wang2023voyager}. Despite these advances, repository migration remains challenging because it requires maintaining global consistency over thousands of interdependent files across many optimization iterations. Moreover, generation and evaluation are often performed by the same agent, making feedback susceptible to self-confirmation bias
\cite{zheng2023judging,huang2024large}. ECAT instead separates generation from evaluation through a generator--discriminator architecture and continuously evolves a shared hierarchical memory that stores transferable migration knowledge.

\paragraph{Code migration.}

Recent advances in large code LLMs have substantially improved code migration and translation capabilities \cite{chen2021codex,roziere2023codellama,li2023starcoder}.
Recent work further extends it from individual functions to entire repositories by explicitly modeling project structures and cross-file dependencies. AlphaTrans performs hierarchical repository translation through static analysis and intermediate validation \cite{ibrahimzada2025alphatrans},
Skeleton-Guided Translation constrains generation with repository skeletons \cite{zhang2025skeleton},
RepoTransBench introduces executable repository benchmarks together with a ReAct-based translation agent
\cite{wang2025repotransbench}, and ReCodeAgent coordinates multiple specialized agents for repository translation \cite{ibrahimzada2026recodeagent}.
The repositories used in RepoTransBench and ReCodeAgent contain approximately 2,394 and 1,975 LOC on average, respectively, making repository reasoning relatively short-horizon.
Android-to-HarmonyOS migration focus primarily on individual migration stages, including UI translation \cite{gong2025uitrans,zheng2026arktrans}, Java-to-ArkTS translation \cite{liu2024java2arkts}, and ArkTS program repair \cite{xie2026arkeval}.
In contrast, our proposed A2H-RepoBench contains real-world Android repositories ranging from 50K to 300K LOC, where migration becomes a long-horizon optimization problem requiring consistent reasoning over hundreds of interdependent source files and platform-specific components.
Rather than translating individual components independently, ECAT jointly optimizes source code, resources, and runtime behaviors through an entropy-guided adversarial optimization loop.

\begin{figure*}[!ht]
    \centering
    \includegraphics[width=1.0\linewidth]{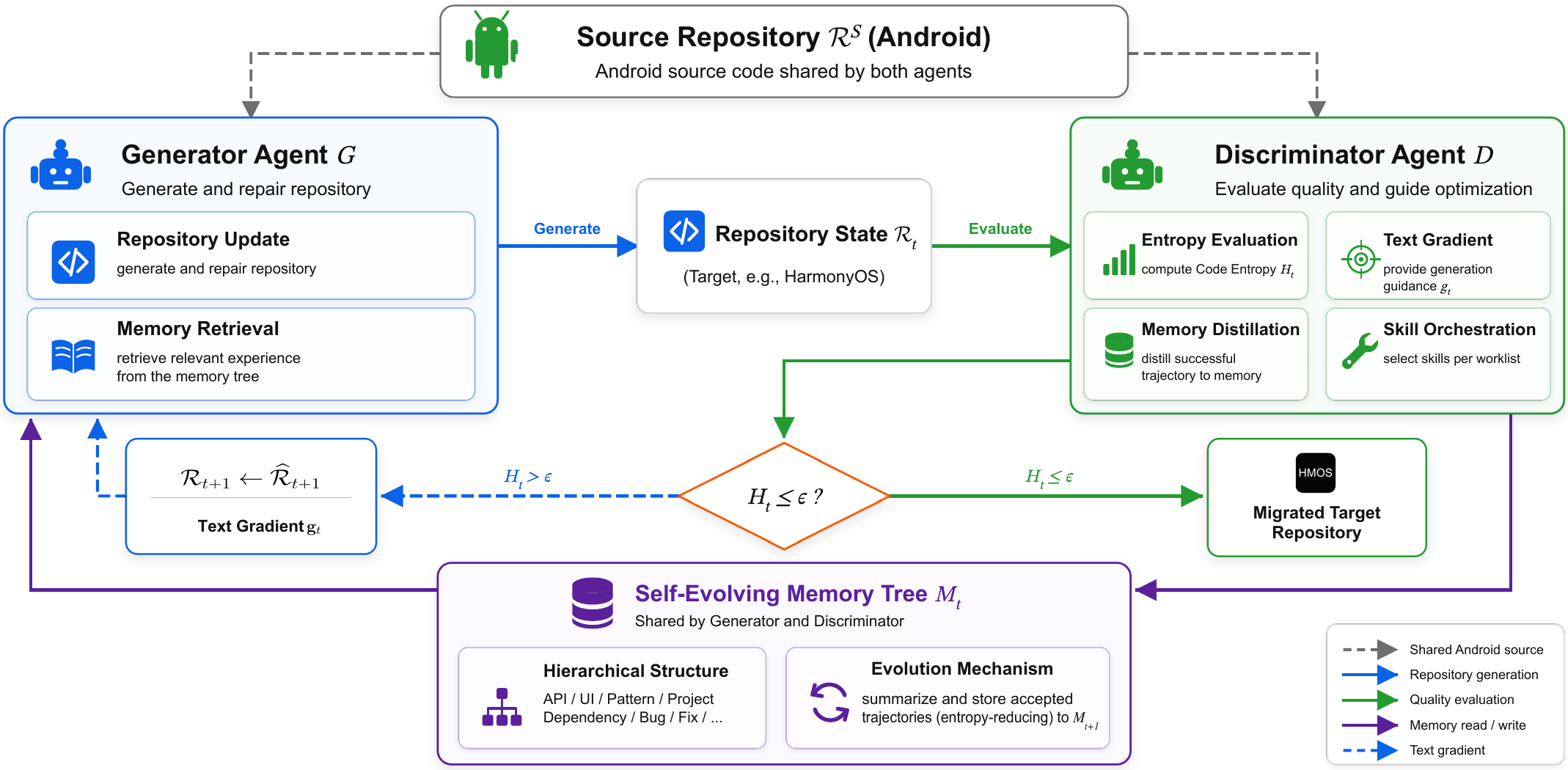}
    \caption{\textbf{Overview of ECAT.} ECAT performs repository migration through an entropy-based adversarial optimization loop between a Generator agent $G$ and an independent Discriminator agent $D$. Starting from a source Android repository, the Generator iteratively updates the target HarmonyOS repository, while the Discriminator estimates Code Entropy and produces text gradients to guide repository optimization. The optimization terminates when the estimated entropy falls below a predefined threshold $\epsilon$. Entropy-reducing updates are distilled into the shared self-evolving memory tree.}
    \label{fig:framework}
\end{figure*}

\section{Method}
\label{sec:methods}
In this section, we present our method of \method{} to migrate software repositories across platforms automatically.

\subsection{Problem Formulation and Method Overview}

Given a source application repository $\mathcal{R}^{S}=\{f_i^{S}\}_{i=1}^{\mathcal{N}}$, where $\mathcal{N}$ denotes the number of source files and $f_i^{S}$ denotes the $i$-th source file, our goal is to automatically migrate it into a target HarmonyOS repository $\mathcal{R}^{T}$. Unlike file-level code translation, repository migration requires preserving consistency across numerous interdependent source files, project configurations, and navigation graphs. As repository size grows, sparse attention in long-context LLMs gradually weakens global dependency modeling, making repository-level consistency increasingly difficult to preserve.

LLM-based repository generation is inherently probabilistic, causing migration errors to accumulate throughout the repository and progressively increase repository-level disorder. We characterize this disorder using the proposed \emph{Code Entropy} $H(\mathcal{R})$, where higher entropy indicates greater uncertainty and unresolved migration errors. Repository migration is therefore formulated as the following entropy minimization problem,

\begin{equation}
\pi^{*}
=
\arg\min_{\pi}
\mathbb{E}\!\left[
H(\mathcal{R}^{T})
\right],
\label{eq:objective}
\end{equation}
where $\pi^{*}$ denotes the optimal migration policy, $\mathcal{R}^{T}$ denotes the final migrated repository after optimization, and the expectation is taken over the stochastic repository generation process.

To solve this repository-level optimization problem, we propose Entropy-based Code Adversarial Translation (ECAT) as shown in Figure~\ref{fig:framework}, which decouples repository generation from repository evaluation through two specialized agents: a generator $G(\cdot)$ and an independent discriminator $D(\cdot)$. At optimization iteration $t$, the discriminator analyzes the current migrated repository and produces an estimated Code Entropy together with a structured text gradient $\mathbf{g}_t$, while the generator updates the repository according to the generated optimization signal. Once the estimated entropy falls below a predefined threshold, the optimization terminates and outputs the final HarmonyOS repository. Successful entropy-reducing optimization trajectories are distilled into the shared self-evolving memory tree $\mathcal{M}_{t}$, enabling both repository generation and repository evaluation to continuously improve across migration tasks without updating model parameters.

\subsection{Repository-level Code Entropy}
\label{sec:entropy}

Repository migration quality cannot be adequately characterized by a single criterion such as compilation success or runtime execution, since a migrated repository may successfully compile while still containing missing functionalities, inconsistent navigation, or platform-specific migration errors. We therefore introduce \emph{Code Entropy}, a unified repository-level optimization objective score that quantifies the overall disorder of a migrated repository. We use entropy as an analogy to characterize repository-level uncertainty caused by unresolved migration inconsistencies, rather than a strict information-theoretic entropy. Code Entropy serves as a computable quality proxy that aggregates migration errors across multiple complementary dimensions into a single optimization objective whose value decreases as migration defects are progressively resolved. Consequently, higher Code Entropy indicates greater uncertainty, inconsistency, and unresolved migration errors, whereas lower Code Entropy corresponds to a structurally complete, semantically faithful, and functionally executable repository.

Given the source Android repository $\mathcal{R}^{S}$ and the migrated HarmonyOS repository state $\mathcal{R}_{t}$, repository quality is characterized from $K$ complementary entropy dimensions. Let

\[
\mathbf{E}
=
[e_1,e_2,\ldots,e_K]^{\top},
\qquad
e_i\in[0,1],
\]
denote the repository entropy vector, where $e_i$ is the normalized entropy score, i.e., the uncertainty of migration correctness for the $i$-th evaluated entropy dimension. Each dimension is assigned a non-negative importance weight

\[
\mathbf{W}
=
[w_1,w_2,\ldots,w_K]^{\top},
\qquad
w_i\ge0.
\]
The weights $w_i$ encode the relative severity of different migration defects and are used to primarily govern repair prioritization. In practice, we assign coarse weights in the range $[0.1,0.3]$. ECAT is insensitive to their exact values, and a sensitivity analysis is provided in the supplementary material. The overall Code Entropy is computed as

\begin{equation}
H(\mathcal{R}_t)
=
\frac{\mathbf{W}^{\top}\mathbf{E}}
{\mathbf{1}^{\top}\mathbf{W}},
\label{eq:entropy}
\end{equation}
where $\mathbf1$ denotes the all-one vector for normalization. Consequently, $H(\mathcal{R})\in[0,1]$, with larger values indicating greater repository disorder.
Each entropy score is estimated as

\begin{equation}
e_i=\psi_i(\mathcal{R}^{S},\mathcal{R}_t),
\label{eq:estimator}
\end{equation}
where $\psi_i(\cdot)$ denotes the entropy estimator of the $i$-th evaluated entropy dimension. Depending on the observability of the migration defect, $\psi_i(\cdot)$ is implemented using either deterministic rule-based verification or agentic LLM judges. Deterministic verification is applied to objectively observable properties such as compilation success and placeholder counts, whereas LLM judges are employed for inherently semantic properties including migration fidelity, runtime behavior, and UI consistency. Notably, all entropy dimensions are defined at the level of generic migration defects and contain no application-specific functional requirements.

For clarity, we divide entropy dimensions into static and dynamic entropy. Static entropy evaluates repository correctness without execution, including compilation, structure, and migration fidelity, whereas dynamic entropy captures runtime defects after deployment on the HarmonyOS emulator, such as crashes and UI inconsistencies. Both categories are normalized and aggregated through Eq.~(\ref{eq:entropy}) to form the unified Code Entropy objective.

\subsection{Entropy-guided Adversarial Optimization}
\label{sec:adversarial}

Code Entropy provides a unified optimization objective for the migration, while the remaining challenge is how to reduce repository disorder during long-horizon translation. Repository migration naturally involves two processes: identifying unresolved defects and repairing them. ECAT therefore employs two specialized agents with distinct responsibilities. Inspired by adversarial verification, the discriminator acts as an adversarial evaluator that challenges the current repository state by exposing high-entropy regions and generating optimization signals, while the generator improves the repository by eliminating the identified defects. 
Both agents are re-instantiated with isolated contexts at every iteration, so neither can access the other's reasoning traces or inherit bias from its own previous judgments.
Their interaction establishes a challenge-and-repair optimization loop that progressively drives repository states toward lower code entropy.

\begin{figure}[!htbp]
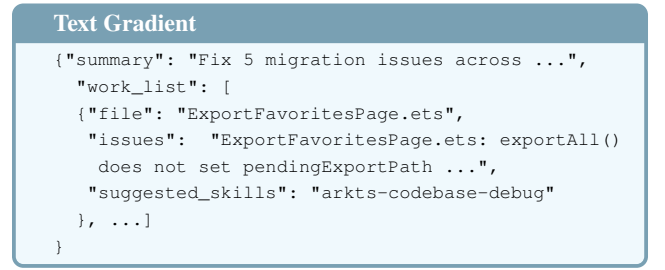

\centering
\begin{tcolorbox}[
    colback=cyan!5,
    colframe=cyan!30!gray,
    title=\small \textbf{Text Gradient},
    fonttitle=\bfseries,
    fontupper=\ttfamily\scriptsize,
    top=1pt, bottom=1pt,
]
\begin{verbatim}
{"summary": "Fix 5 migration issues across ...",
  "work_list": [
  {"file": "ExportFavoritesPage.ets",
   "issues":  "ExportFavoritesPage.ets: exportAll() 
    does not set pendingExportPath ...",
   "suggested_skills": "arkts-codebase-debug"
  }, ...] 
}
\end{verbatim}
\end{tcolorbox}
\caption{An example of the structured text gradient generated by the discriminator in the Gallery repository migration. }
\label{fig:text_gradient}
\end{figure}

At optimization iteration $t$, given the source repository $\mathcal{R}^{S}$, the current migrated repository $\mathcal{R}_{t}$, and the shared self-evolving memory tree $\mathcal{M}_{t}$, the discriminator evaluates the current repository and produces an estimated Code Entropy together with a structured text gradient,

\begin{equation}
(H_t,\mathbf{g}_t)
=
D(\mathcal{M}_{t};
\mathcal{R}^{S},
\mathcal{R}_{t}),
\label{eq:discriminator}
\end{equation}
where $H_t$ denotes the estimated Code Entropy of the current repository and $\mathbf g_t$ denotes the corresponding text gradient for repository optimization.
Because repository optimization is performed in the discrete code space, numerical gradients cannot be propagated through repository states. Instead, the discriminator translates repository defects into structured text gradients \cite{yuksekgonul2024textgrad} that explicitly identify high-entropy regions, diagnose migration failures, prioritize repair targets, and recommend the corresponding skills as shown in Figure \ref{fig:text_gradient}.

Guided by the exposed defects, the generator updates the repository according to

\begin{equation}
\mathcal R_{t+1}
=
G
\!\left(
\mathbf g_t,
\mathcal M_t;
\mathcal R^{S},
\mathcal R_t
\right),
\label{eq:transition}
\end{equation}
where $\mathbf g_t$ denotes the structured text gradient generated by the discriminator and $\mathcal M_t$ denotes the shared self-evolving memory. These structured optimization signals enable the generator to coordinate edits across multiple correlated source files, project configurations, resource files, and navigation graphs while preserving repository-level consistency.
At each iteration, the discriminator challenges the current repository by exposing previously unresolved defects, while the generator repairs these defects to reduce Code Entropy. This alternating process progressively drives the repository toward lower-disorder states.

To guarantee entropy reduction, a candidate repository is accepted only if its Code Entropy decreases. Let $\widehat{\mathcal R}_{t+1}$ denote the candidate repository generated after applying the current text gradient. The accepted repository is determined by

\begin{equation}
\mathcal R_{t+1}
=
\begin{cases}
\widehat{\mathcal R}_{t+1},
&
H(\widehat{\mathcal R}_{t+1})
<
H(\mathcal R_t),
\\
\mathcal R_t,
&
\text{otherwise},
\end{cases}
\label{eq:accept}
\end{equation}
which guarantees that every accepted repository update monotonically reduces estimated Code Entropy.
To avoid premature convergence caused by transient entropy fluctuations, ECAT adopts a sliding-window stopping criterion. Optimization terminates when the repository Code Entropy remains below a predefined threshold $\epsilon$ during the most recent $l$ iterations,

\begin{equation}
H(\mathcal R_{t-i}) \le \epsilon,
\qquad
i=0,\ldots,l-1,
\label{eq:stop}
\end{equation}
where $l$ denotes the sliding-window size.

\subsection{Self-evolving Memory Tree}
\label{sec:memory}

\begin{figure}[!ht]
    \centering
    \includegraphics[width=\linewidth]{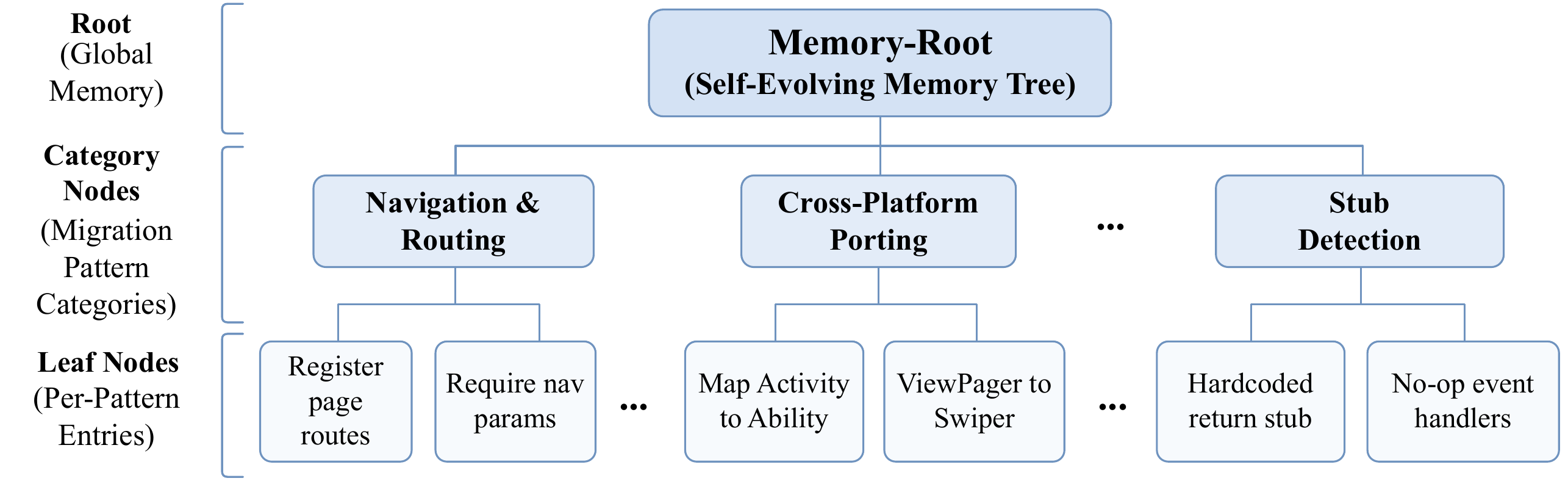}
    \caption{The hierarchical self-evolving memory tree.}
    \label{fig:memory}
\end{figure}

Unlike prior agent memory that stores flat experiences \cite{packer2023memgpt}, continual improvement is achieved through a shared self-evolving memory tree, which serves as the only learnable component continuously updated during deployment. The memory accumulates reusable migration experience from entropy-reducing optimization trajectories and provides transferable knowledge for repository generation. As migration experience grows, a flat memory becomes increasingly inefficient and introduces excessive retrieval context. ECAT therefore organizes the accumulated experience hierarchically, enabling coarse-to-fine retrieval that focuses only on defect-relevant branches.
Figure~\ref{fig:memory} illustrates the memory organization. Each layer of the tree has an index directory for the agent to automatically retrieve relevant memories layer by layer. The root node stores global migration knowledge, intermediate nodes organize reusable patterns by coarse defect categories (e.g., navigation adaptation, cross-platform porting, and stub detection), and leaf nodes maintain generalized migration patterns consisting of representative defects, root causes, and verification criteria. Repository-specific implementation details are discarded to maximize transferability across repositories.

\begin{algorithm}[!ht]
\caption{Entropy-based Code Adversarial Translation}
\label{alg:ecat}
\begin{algorithmic}[1]

\REQUIRE Source repository $\mathcal{R}^{S}$, initial target repository $\mathcal{R}_{0}$, initial memory tree $\mathcal{M}_{0}$, maximum iterations $\tau$, entropy threshold $\epsilon$
\ENSURE Migrated target repository $\mathcal{R}^{T}$

\STATE $(H_{0},\mathbf{g}_{0})
\leftarrow
D(\mathcal{M}_{0};\mathcal{R}^{S},\mathcal{R}_{0})$

\FOR{$t=0$ \TO $\tau-1$}

    \IF{$H_{t} \le \epsilon$}
        \RETURN $\mathcal R_t$
    \ENDIF

    \STATE $\widehat{\mathcal{R}}_{t+1}
    \leftarrow
    G(\mathbf{g}_{t},
    \mathcal{M}_{t};
    \mathcal{R}^{S},
    \mathcal{R}_{t})$

    \STATE $(H_{t+1},\mathbf{g}_{t+1})
    \leftarrow
    D(\mathcal{M}_{t};
    \mathcal{R}^{S},
    \widehat{\mathcal{R}}_{t+1})$

    \IF{$H_{t+1}<H_{t}$}

        \STATE $\mathcal{R}_{t+1}
        \leftarrow
        \widehat{\mathcal{R}}_{t+1}$,
        Distill to memory $\mathcal{M}_{t+1}$.

    \ELSE

        \STATE $\mathcal{R}_{t+1}
        \leftarrow
        \mathcal{R}_{t}$, 
        $\mathcal{M}_{t+1}
        \leftarrow
        \mathcal{M}_{t}$

    \ENDIF

\ENDFOR

\STATE $\mathcal{R}^{T}
\leftarrow
\mathcal{R}_{t+1}$

\RETURN $\mathcal{R}^{T}$

\end{algorithmic}
\end{algorithm}

After each optimization iteration, the discriminator distills successful entropy-reducing trajectories into the corresponding memory branches. Newly discovered patterns are merged with existing nodes whenever possible, while redundant entries are consolidated to maintain a compact hierarchy. During migration, both agents retrieve task-relevant branches to guide repository repair, defect localization, entropy estimation, and text-gradient generation.   Consequently, ECAT self-evolves by continually acquiring and organizing transferable migration knowledge in the memory tree, enabling sustained performance improvements across repositories.
The overall ECAT procedure is summarized in Algorithm~\ref{alg:ecat}.

\section{Experiments}

\subsection{Benchmark and Metrics}

To evaluate repository-level Android-to-HarmonyOS migration, we construct \textbf{A2H-RepoBench}, a benchmark consisting of three representative open-source Android repositories spanning different repository scales. Unlike existing benchmarks that primarily focus on file-level translation, A2H-RepoBench emphasizes repository-level migration, requiring models to simultaneously preserve cross-file dependencies, resource consistency, and runtime functionality throughout long-horizon repository optimization.
The benchmark contains three real-world applications with progressively increasing repository complexity: \emph{Gallery (50K LOC)} \cite{gallery}, \emph{AntennaPod (120K LOC)} \cite{antennapod}, and \emph{Meshtastic (300K LOC)} \cite{meshtastic}. Gallery is a multimedia application that mainly evaluates user interface migration and media management. AntennaPod is a podcast management application involving multimedia playback and asynchronous task scheduling, presenting substantially more complex cross-module dependencies. Meshtastic is a large-scale mesh communication application supporting Bluetooth communication and hardware interaction, posing significant challenges for long-range dependency preservation. Together, these repositories cover repository scales ranging from tens of thousands to hundreds of thousands of lines of code, enabling a comprehensive evaluation under increasingly challenging migration scenarios.

Following migration, repository quality is evaluated from two perspectives: \emph{Node Alignment} and \emph{Agent-as-Judge} \cite{zhuge2025agent}. Node Alignment measures structural preservation by constructing platform-agnostic semantic graphs for the Android and HarmonyOS repositories. Specifically, semantic graphs for both the Android and HarmonyOS repositories are constructed using CodeGraph~\cite{codegraph}, and node alignment is computed by matching their platform-agnostic semantic nodes. The alignment score is then computed as

\begin{equation}
\mathrm{Align}
=
\frac{|M|}{|V_s|},
\end{equation}
where $V_s$ denotes the semantic nodes of the Android repository and $M$ denotes the matched node pairs in HarmonyOS. Agent-as-Judge evaluates functional preservation using a predefined feature checklist, where each feature is assigned a score $s_i$ of $1$, $0.5$, or $0$ (\textsc{Full}, \textsc{Partial}, or \textsc{Missing}), yielding

\begin{equation}
\mathrm{Agent}
=
\frac{1}{N}\sum_{i=1}^{N} s_i,
\end{equation}
where $N$ is the number of evaluated features. Note that the discriminator in Sec. \ref{sec:methods} evaluates only application-agnostic defect categories, with no access to the feature checklist used at evaluation. To facilitate reproducible evaluation, we package the entire protocol as a reusable skill, \texttt{a2h-evaluate}, which can be directly invoked by arbitrary LLM agents. 
More details are provided in the supplementary material.

\subsection{Implementation Details and Baselines}

Unless otherwise specified, all methods use DeepSeek-V4-Pro as the underlying LLM, while the second-stage dynamic Code Entropy optimization uses the multimodal LLM Qwen3.7-Plus to evaluate runtime GUI states from emulator screenshots. A three-level self-evolving memory tree is shared by the generator and discriminator. Repository optimization terminates once the estimated Code Entropy remains below a predefined threshold within a sliding window of size two. 
We fix $\epsilon=0.01$ for all experiments.
All methods are evaluated under identical execution environments and repository inputs.

We compare ECAT with three representative agentic software engineering frameworks. OpenHands~\cite{openhands2024} is a general-purpose software engineering agent that performs repository understanding, code editing, command execution, and tool invocation through a ReAct-style self-feedback interaction loop. RepoTransAgent~\cite{wang2025repotransbench} is a ReAct-based repository translation agent that iteratively alternates between repository inspection, code editing, command execution, and repository search in a self-feedback manner. ReCodeAgent~\cite{ibrahimzada2026recodeagent} is a multi-agent repository translation framework that coordinates specialized agents for repository analysis, planning, code translation, and validation. 


\begin{table}[!ht]
\centering
\caption{Comparison results (\%) of different methods across code repositories in A2H-RepoBench. ``w/o DE'' means removing emulator-based Dynamic Entropy evaluation.}
\label{tab:main_results}

\setlength{\tabcolsep}{3pt}
\renewcommand{\arraystretch}{1.05}
\resizebox{\linewidth}{!}
{%
\begin{tabular}{l|cc|cc|cc|c}
\toprule
\multirow{2}{*}{\textbf{Method}} &
\multicolumn{2}{c|}{\textbf{Gallery}} &
\multicolumn{2}{c|}{\textbf{AntennaPod}} &
\multicolumn{2}{c|}{\textbf{Meshtastic}} &
\multirow{2}{*}{\textbf{Avg.}} \\
\cmidrule(lr){2-3}
\cmidrule(lr){4-5}
\cmidrule(lr){6-7}
& Align & Agent
& Align & Agent
& Align & Agent
& \\
\midrule

OpenHands
& 40.0 & 4.3
& 9.8 & 12.2
& 8.1 & 9.7
& 14.0 \\

RepoTransAgent
& 36.0 & 10.5
& 4.2 & 20.4
& 2.3 & 6.6
& 13.3 \\

ReCodeAgent
& 84.0 & 36.1
& 44.0 & 21.7
& 59.9 & 32.6
& 46.4 \\

\midrule

\textbf{ECAT} w/o DE
& {\textbf{85.3}} & {83.4}
& {62.4} & {\textbf{74.7}}
& {61.5} & {68.6}
& {72.7} \\

\textbf{ECAT} (Ours)
& {84.9} & \textbf{85.6}
& \textbf{69.5} & \textbf{74.7}
& \textbf{62.0} & \textbf{71.2}
& \textbf{74.7} \\

\quad
& $\pm$ {\footnotesize 0.8} & $\pm$ {\footnotesize 4.9}
& $\pm$ {\footnotesize 6.5} & $\pm$ {\footnotesize 2.7}
& $\pm$ {\footnotesize 0.6} & $\pm$ {\footnotesize 2.3}
& $\pm$ {\footnotesize 1.7} \\

\bottomrule
\end{tabular}%
}
\end{table}

\begin{figure*}[!ht]
    \centering
    \includegraphics[width=\linewidth]{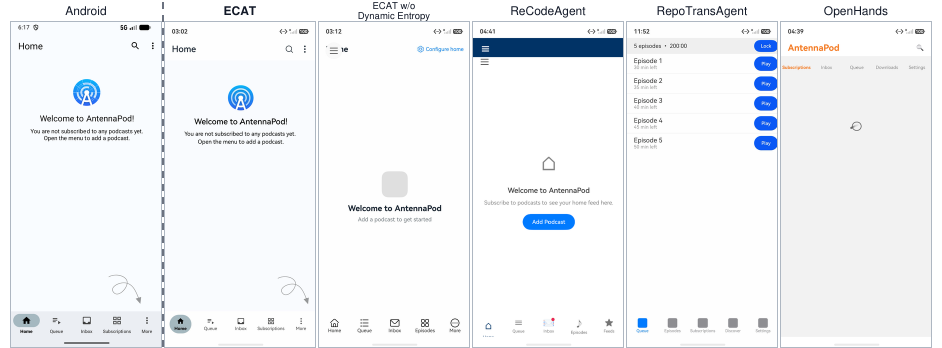}
    \caption{Qualitative comparison of the migrated AntennaPod home page. ECAT preserves the original Android layout, navigation structure, and functional widgets with high visual fidelity. 
    }
    \label{fig:ui_comparison}
\end{figure*}

\subsection{Results}
\paragraph{Quantitative Results.}
Table~\ref{tab:main_results} reports the quantitative results on three real-world repositories of varying scales. All ECAT results are averaged over three independent runs, with standard deviations reported in the last row. ECAT consistently achieves the best performance under both Node Alignment and Agent-as-Judge across all repositories, improving the average score from 46.4\% achieved by the strongest baseline to 74.7\%.
Notably, OpenHands and RepoTransAgent suffer from self-evaluation bias, while ReCodeAgent, though validated independently, lacks an iterative repair loop.
ReCodeAgent attains relatively high Node Alignment but much lower Agent-as-Judge scores because many translated classes preserve the repository structure while containing only shallow implementations or placeholder logic. Although larger repositories such as Meshtastic remain more challenging due to extensive cross-file dependencies, ECAT consistently maintains the best performance under both metrics.  
Removing Dynamic Entropy drops the average from 74.7\% to 72.7\%, mainly on runtime-heavy repositories, while Gallery remains barely affected.
These results demonstrate that decoupling repository generation from quality evaluation and iteratively minimizing Code Entropy leads to more faithful repository migration than existing methods.
Additional results including different base LLMs are provided in the supplementary.

\begin{figure*}[!ht]
    \centering
    \includegraphics[width=0.49\linewidth]{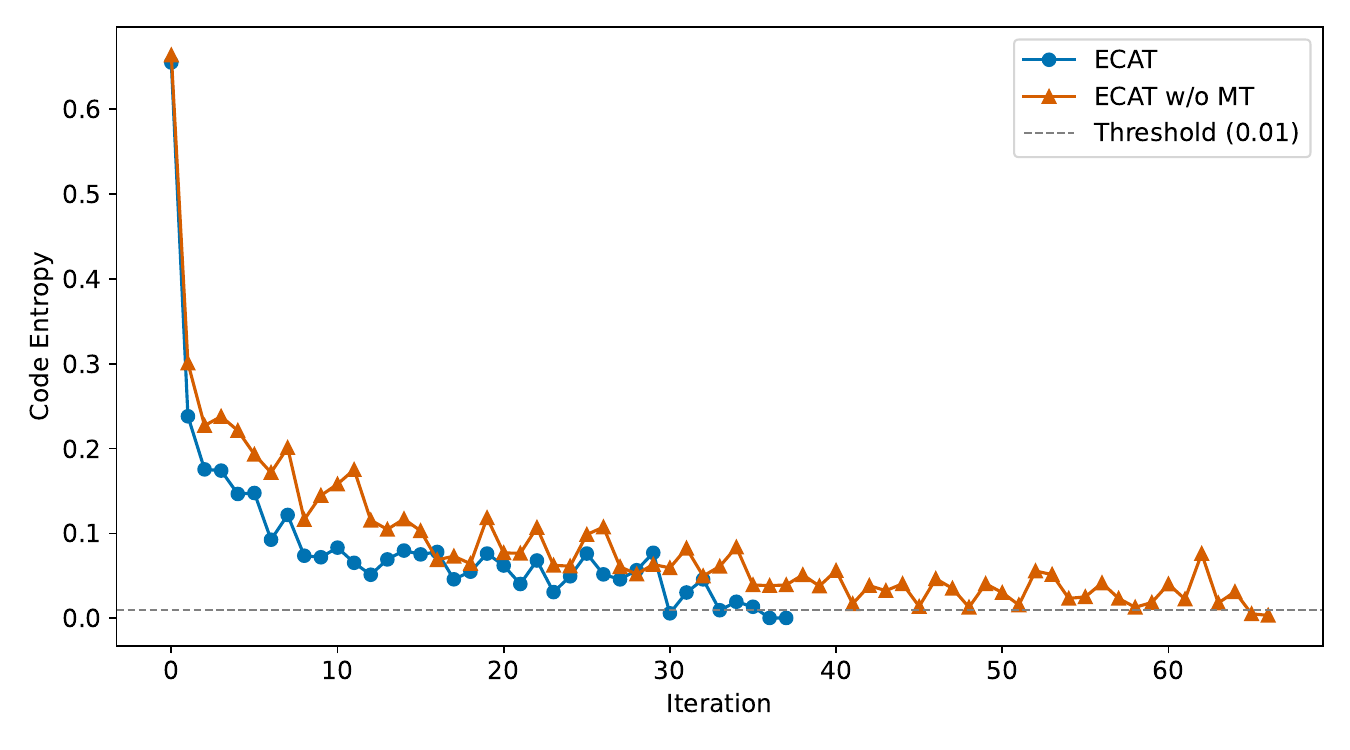}
    \hfill
    \includegraphics[width=0.49\linewidth]{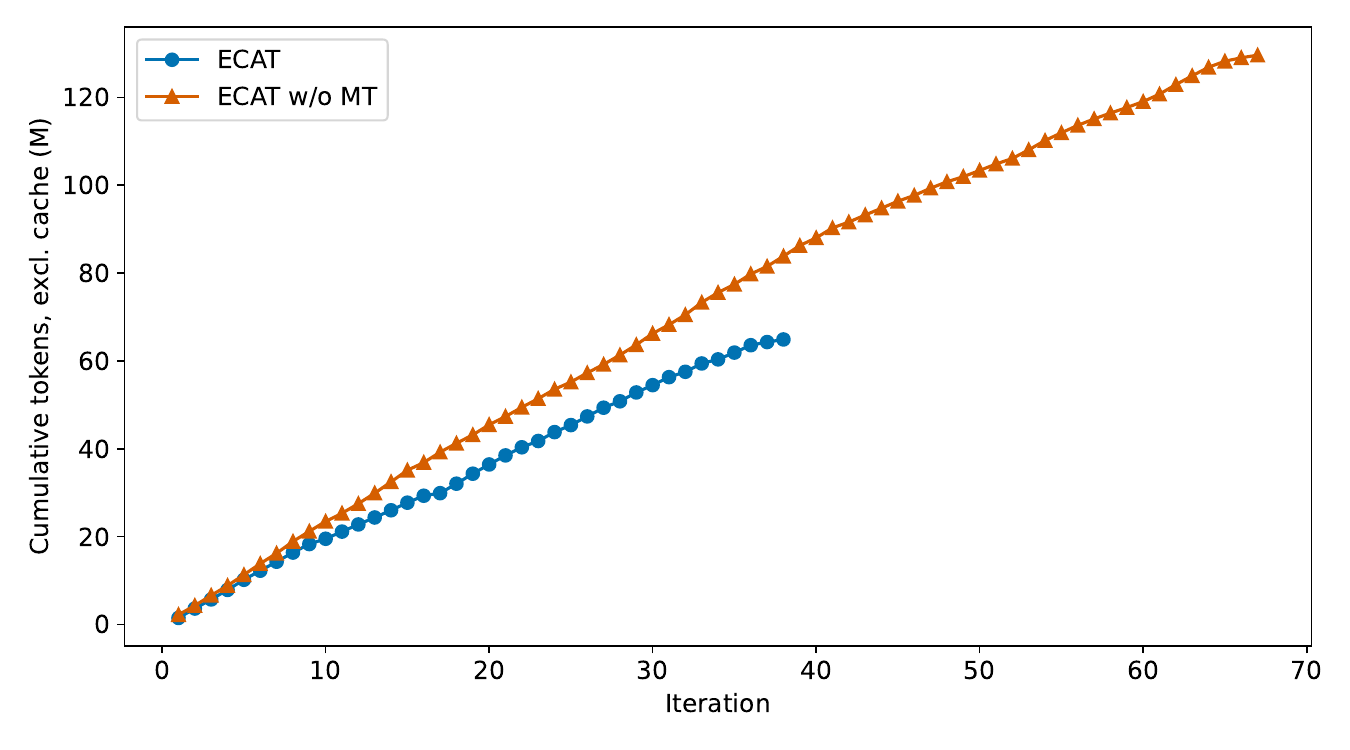}
    \caption{
    Optimization dynamics of ECAT on the \emph{Gallery} repository.
    \textbf{(a)} Evolution of Code Entropy during adversarial optimization. A memory tree (MT) built from successful migration trajectories of the other repositories enables faster entropy reduction and earlier convergence.
    \textbf{(b)} Corresponding cumulative token consumption. Faster convergence with the memory tree substantially reduces inference cost while reaching the same entropy threshold.
    }
    \label{fig:entropy}
\end{figure*}

\paragraph{Qualitative UI Comparison.}
Figure~\ref{fig:ui_comparison} presents a qualitative comparison of the migrated home page of the AntennaPod application. ECAT produces a HarmonyOS interface that closely matches the original Android application in both layout and functionality, preserving the navigation bar and interactive widgets. Removing Dynamic Entropy leads to noticeable UI degradation, including missing icons and incomplete layout rendering, indicating that static verification alone cannot adequately capture runtime interface quality. Existing repository migration agents exhibit more severe failures.  These qualitative results demonstrate that minimizing Code Entropy through adversarial optimization substantially improves both structural fidelity and functional completeness of the migrated repository. UI comparison for the other repositories is provided in the supplementary.

\paragraph{Optimization dynamics and memory transfer.}
Figure~\ref{fig:entropy} illustrates the optimization process of ECAT on the \emph{Gallery} repository. The self-evolving memory tree is initialized exclusively from successful migration trajectories of the other repositories in A2H-RepoBench to evaluate cross-repository transfer. As shown in Figure ~\ref{fig:entropy}(a), Code Entropy decreases rapidly during the early iterations as major migration defects are repaired, and then gradually converges as the remaining errors become increasingly fine-grained. With the memory tree, ECAT reaches the stopping criterion in only 38 iterations, compared with 66 iterations without memory (w/o MT), demonstrating that transferable migration knowledge accelerates defect localization and repair. This faster convergence also substantially reduces inference cost. As shown in Figure ~\ref{fig:entropy}(b), ECAT with memory consumes nearly half as many cumulative tokens ($\sim$65M vs. $\sim$125M) while achieving the same stopping criterion, demonstrating that the proposed self-evolving memory tree improves both optimization efficiency and cross-repository knowledge transfer.

\section{Conclusion}

This paper presented \method{}, an entropy-guided adversarial framework for long-horizon Android-to-HarmonyOS repository migration. By formulating repository migration as an entropy minimization problem, ECAT iteratively reduces repository disorder through adversarial interaction between an independent generator and discriminator, while continuously improving via a self-evolving memory tree that accumulates transferable migration knowledge. We also introduce A2H-RepoBench, the first benchmark for long-horizon Android-to-HarmonyOS repository migration, together with functional specifications and a unified evaluation protocol. Extensive experiments demonstrate that ECAT consistently outperforms existing LLM-based agentic methods on large-scale real-world repositories.

Despite these promising results, ECAT still has several limitations. The iterative adversarial optimization inevitably increases inference latency and token consumption compared with one-shot migration. 
Future work will investigate more efficient optimization strategies and the extension of ECAT to broader repository-level software engineering tasks. 

\bibliography{aaai2027}


\appendix

\section{Additional Results}

\subsection{Effect of the Independent Discriminator}
To directly verify that decoupling generation from evaluation is essential, we evaluate a w/o Discriminator variant in which a single agent both updates and evaluates the repository within a shared context. This variant retains the full skill library. As shown in Table~\ref{tab:ablation_disc}, removing the independent discriminator collapses the average score (unweighted over the three repositories) from 74.7\% to 28.4\%, a larger degradation than removing any other component, and the failure amplifies with repository scale. On Gallery, the smallest repository, the variant still completes 104 of 323 features in full, but both metrics drop by roughly 40 points as a third of the features are left missing. On AntennaPod, the degradation shifts from missing features to shallow ones: 106 of 184 features are judged Partial while only 5 remain Full, indicating that the agent touches most features but stops before completing them. On Meshtastic, the largest repository, the variant collapses almost entirely, with only 7 of 129 features fully functional and 105 missing. Inspecting the optimization trajectories, we observe that the self-evaluating agent systematically overestimates the quality of its own outputs and triggers the stopping criterion prematurely, which explains this scale-dependent pattern: on small repositories premature stopping leaves a partial migration, whereas on large repositories it halts long before the long-horizon optimization can cover the feature space. These results provide direct evidence for the self-confirmation bias: without an independent evaluator, the challenge-and-repair loop degenerates into self-refinement and loses the ability to sustain long-horizon optimization.

\begin{table}[!ht]
\centering
\small
\begin{tabular}{@{}llccc@{}}
\toprule
Repository & Method & Align & Agent & F/P/M \\
\midrule
\multirow{2}{*}{Gallery} & ECAT & \textbf{84.9} & \textbf{85.6} & 256/42/26 \\
 & w/o Discriminator & 46.7 & 45.5 & 104/86/133 \\
\midrule
\multirow{2}{*}{AntennaPod} & ECAT & \textbf{69.5} & \textbf{74.7} & 125/26/34 \\
 & w/o Discriminator & 23.7 & 31.5 & 5/106/73 \\
\midrule
\multirow{2}{*}{Meshtastic} & ECAT & \textbf{62.0} & \textbf{71.2} & 79/26/24 \\
 & w/o Discriminator & 10.7 & 12.0 & 7/17/105 \\
\bottomrule
\end{tabular}
\caption{Ablation of the independent discriminator. In the w/o Discriminator variant, a single agent both generates and evaluates the repository within a shared context, with access to the \textbf{same skill library} as ECAT. F/P/M denotes the number of features judged Full/Partial/Missing by Agent-as-Judge.}
\label{tab:ablation_disc}
\end{table}

\begin{table}[!ht]
\centering
\begin{tabular}{@{}lrrrr@{}}
\toprule
Repository & LOC & Iterations & Time (h) & Tokens (M) \\
\midrule
Gallery     & 50K  & 38  & $\sim$12 & $\sim$65  \\
AntennaPod  & 120K & 95  & $\sim$29 & $\sim$180 \\
Meshtastic  & 300K & 126 & $\sim$44 & $\sim$230 \\
\bottomrule
\end{tabular}
\caption{Migration cost of ECAT across repositories using DeepSeek-V4-Pro. Tokens are cumulative and exclude cache hits; wall-clock time includes emulator-based dynamic entropy evaluation.}
\label{tab:cost_repo}
\end{table}

\begin{table}[!ht]
\centering
\small
\begin{tabular}{@{}lccccc@{}}
\toprule
Base LLM & Align & Agent & Iters. & Time (h) & Tokens (M) \\
\midrule
DS-V4-Pro   & 84.9 & 85.6 & 38 & $\sim$12 & $\sim$65  \\
DS-V4-Flash & 78.7 & 87.1 & 89 & $\sim$24 & $\sim$146 \\
GLM-5.2           & {86.7} & {88.1} & {29} & $\sim$27 & {$\sim$46} \\
\bottomrule
\end{tabular}
\caption{ECAT with different base LLMs on the Gallery repository. Quality (\%) is evaluated by Node Alignment and Agent-as-Judge; cost is reported as iterations, wall-clock time, and cumulative tokens. ``DS'' means DeepSeek.}
\label{tab:cost_llm}
\end{table}

\begin{table*}[!ht]
\centering
\begin{tabular}{llp{8.6cm}cc}
\toprule
Category & Dimension & Description & Estimator & $w_i$ \\
\midrule
\multirow{12}{*}{Static}
 & \texttt{compile} & Whether the migrated repository compiles & Rule & 0.30 \\
 & \texttt{static\_lint} & Untranslated Android idioms and ArkTS anti-patterns & Rule & 0.10 \\
 & \texttt{style} & Coding-style violations (naming, \texttt{console}, \texttt{px}, \texttt{var}, etc.) & Rule & 0.10 \\
 & \texttt{placeholder} & Unfinished stubs: TODOs, placeholders, template text & Rule & 0.10 \\
 & \texttt{app\_identity} & Package name, vendor, version, and icon left as template defaults & Rule & 0.10 \\
 & \texttt{data\_layer\_parity} & Data-layer completeness against Android database tables, entities, and settings keys & Rule & 0.30 \\
 & \texttt{permission\_parity} & Each Android permission mapped to its HarmonyOS equivalent & Rule & 0.30 \\
 & \texttt{page\_audit} & Per-page shell detection via code volume and interaction density & LLM & 0.10 \\
 & \texttt{feature\_check} & Per-feature completeness against the Android counterpart & LLM & 0.30 \\
 & \texttt{hallucination} & Fabricated functionality absent from the Android source & LLM & 0.10 \\
 & \texttt{entry\_nav} & Navigation reachability: unregistered routes, dead tabs, empty entries & LLM & 0.30 \\
 & \texttt{source\_parity} & File-level 1:1 accounting of Android sources, exposing missing files and fake implementations & LLM & 0.10 \\
\midrule
\multirow{2}{*}{Dynamic}
 & \texttt{ui\_align} & Screenshot-level UI consistency scored by a multimodal judge & LLM & 0.10 \\
 & \texttt{runtime\_liveness} & Cold-start liveness and smoke-click probing on the emulator & Rule & 0.10 \\
\bottomrule
\end{tabular}
\caption{Entropy dimensions of Code Entropy. Each dimension $e_i$ is estimated by a dedicated verifier and aggregated via Eq.~(2) with weight $w_i$. All dimensions are defined at the level of generic migration defects against the Android source repository and contain no application-specific functional requirements. ``Rule'' denotes deterministic rule-based verification; ``LLM'' denotes agentic (multimodal) LLM judging.}
\label{tab:entropy_dims}
\end{table*}

\subsection{Migration Cost and Base-LLM Analysis}
Table~\ref{tab:cost_repo} reports the cost of ECAT across the three repositories. Iterations, wall-clock time, and token consumption all grow with repository scale, roughly tripling from Gallery (50K LOC) to Meshtastic (300K LOC), which reflects the long-horizon nature of the task rather than a limitation specific to ECAT: each additional 100K LOC introduces more cross-file dependencies for the discriminator to audit and more defects for the generator to repair. Table~\ref{tab:cost_llm} further evaluates ECAT with different base LLMs on Gallery. ECAT is robust to the choice of base model: all three LLMs reach comparable final quality (83--89\% Agent), indicating that the adversarial entropy-minimization loop, rather than the raw capability of a single model, drives migration quality. The base model instead determines the cost profile, where per-step reliability and inference speed trade off. The weaker but fastest DeepSeek-V4-Flash requires $2.3\times$ more iterations and tokens than DeepSeek-V4-Pro, as each individual repair is less reliable and more defects survive to subsequent rounds; nevertheless, it still converges to slightly higher functional quality, showing that the acceptance criterion in Eq.~(6) compensates for weaker per-step generation through more optimization rounds. Conversely, GLM-5.2 produces the most reliable per-step repairs, converging in the fewest iterations and tokens with the best overall quality, but its slowest inference speed yields the highest wall-clock time despite the fewest rounds.

\subsection{Entropy Dimensions and Weights} 
Table~\ref{tab:entropy_dims} details the $K{=}14$ entropy dimensions, grouped into static entropy (evaluated without execution) and dynamic entropy (runtime behavior on the HarmonyOS emulator). All dimensions measure generic migration defects against the Android source repository and involve no application-specific feature checklist, consistent with the isolation stated in Sec.~4.1. We assign $w_i{=}0.3$ to five dimensions whose defects block or silently break entire features (\texttt{compile}, \texttt{entry\_nav}, \texttt{feature\_check}, \texttt{data\_layer\_parity}, \texttt{permission\_parity}) and $w_i{=}0.1$ to the rest. The weights primarily govern repair prioritization in the text gradient and they also rescale the aggregated entropy in Eq.~(2), so the stopping threshold $\epsilon$ should be calibrated jointly with the weight configuration. ECAT is insensitive to the exact weight values: setting all weights uniformly to 0.1 on Gallery yields 73.3\% Align and 83.4\% Agent, close to the default configuration (84.9\% / 85.6\%), with functional quality largely preserved.

\begin{figure*}[!ht]
\centering
\includegraphics[width=\textwidth]{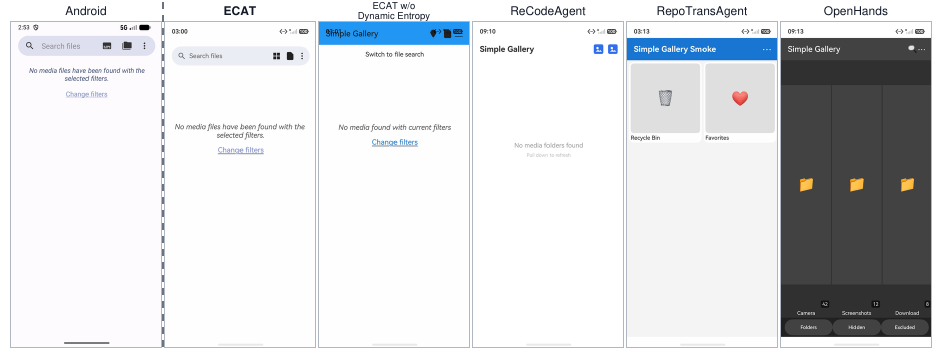}
\par\smallskip
\includegraphics[width=\textwidth]{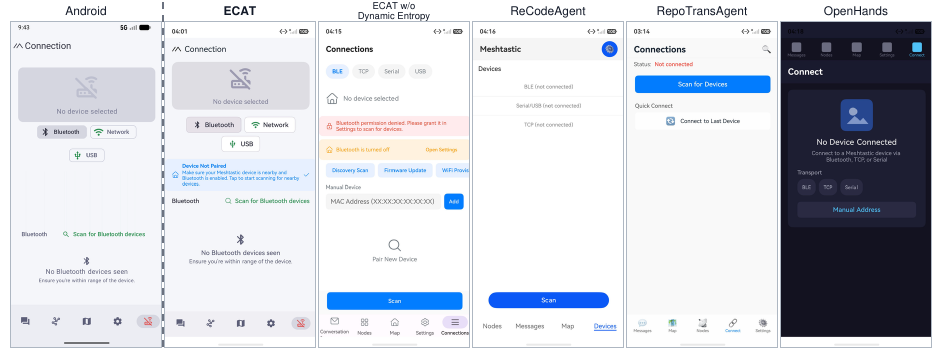}
\caption{Qualitative UI comparisons. Top: Gallery in the same empty-media state. Bottom: Meshtastic in the disconnected-device state. In both panels, columns show Android, ECAT, ECAT w/o Dynamic Entropy, ReCodeAgent, RepoTransAgent, and OpenHands from left to right. ECAT retains the principal source-visible controls and entry states, while the baseline screens omit, simplify, or reconnect several of them.}
\label{fig:ui-comparisons}
\end{figure*}

\subsection{Additional Qualitative UI Comparisons}
\label{sec:qualitative-ui}
AntennaPod is shown in the main paper. Figure~\ref{fig:ui-comparisons} groups the remaining Gallery and Meshtastic comparisons. To ensure comparability, all methods are captured under the same application state. Gallery is shown in the empty-media state and Meshtastic in the disconnected-device state, which are the entry screens a user encounters on first launch without granted media access or a paired device.

On Gallery, ECAT reproduces the Android entry state, including the search bar, the view-mode controls, the empty-state message, and the actionable ``Change filters'' link. The w/o Dynamic Entropy variant preserves the overall layout but exhibits rendering defects invisible to static verification, such as a broken title bar. Among the baselines, ReCodeAgent renders a structurally plausible but simplified shell. RepoTransAgent reconnects the entry to a folder grid that does not correspond to the Android empty state. OpenHands falls back to a generic placeholder layout with misrendered thumbnails.
On Meshtastic, ECAT preserves the connection workflow of the Android original. The transport tabs for Bluetooth, Network, and USB, the no-device placeholder, the scan affordance, and the bottom navigation are all present and correctly wired. The w/o Dynamic Entropy variant retains most controls but introduces runtime-only defects, including misplaced permission banners and an unstyled scan section. The baselines degrade more severely. ReCodeAgent and RepoTransAgent collapse the transport tabs into a single scan button, and OpenHands renders a static connect card whose controls are not reachable from the migrated navigation.

Consistent with the AntennaPod comparison in the main paper, the gap between ECAT and its w/o Dynamic Entropy variant comes mainly from rendering and interaction defects rather than missing features. This explains why the static Agent-as-Judge scores understate the practical impact of dynamic entropy. These screenshots are qualitative evidence recorded separately from Agent-as-Judge and Node Alignment and do not change either score.

\subsection{Generated HarmonyOS Artifact Size}

Table~\ref{tab:artifact-size} compares the size of the final HarmonyOS/ArkTS repositories. ``ETS files'' counts \texttt{.ets} source files and ``ETS lines'' sums their physical lines. These are descriptive statistics that contribute to neither reported metric, but they corroborate the failure modes discussed in the main paper. OpenHands and RepoTransAgent produce only 2K--6K lines regardless of source scale, consistent with self-terminating after migrating a thin shell. ReCodeAgent generates far more code, even exceeding ECAT in file count on Gallery and Meshtastic, yet scores much lower on Agent-as-Judge, indicating that its volume is inflated by structurally faithful but shallowly implemented classes. ECAT is the only method whose artifact size scales with the source repository, from 34K lines on Gallery to 173K on Meshtastic.

\begin{table}[!ht]
\centering
\footnotesize
\begingroup
\setlength{\tabcolsep}{2pt}
\begin{tabular}{@{}llrr@{}}
\toprule
\textbf{Repository} & \textbf{Method} &
\textbf{ETS files} & \textbf{ETS lines} \\
\midrule
\multirow{4}{*}{Gallery} & ECAT & 82 & 33,930 \\
& OpenHands & 27 & 2,568 \\
& RepoTransAgent & 22 & 5,625 \\
& ReCodeAgent & 108 & 9,468 \\
\midrule
\multirow{4}{*}{AntennaPod} & ECAT & 423 & 94,569 \\
& OpenHands & 25 & 4,582 \\
& RepoTransAgent & 14 & 3,799 \\
& ReCodeAgent & 315 & 58,124 \\
\midrule
\multirow{4}{*}{Meshtastic} & ECAT & 464 & 172,661 \\
& OpenHands & 101 & 6,380 \\
& RepoTransAgent & 12 & 4,174 \\
& ReCodeAgent & 777 & 111,952 \\
\bottomrule
\end{tabular}
\endgroup
\caption{Size of the generated HarmonyOS/ArkTS artifacts by repository and method.}
\label{tab:artifact-size}
\end{table}

\section{Reproducibility Protocol}

\subsection{Migration Inputs and Isolation}

Each baseline adapter exposes a read-only Android repository and initializes a writable target workspace from a minimal HarmonyOS/ArkTS project scaffold. The scaffold provides project and module configuration, an entry ability, and a placeholder page, but no migrated application behavior. The Android repository is the sole source of application behavior, and the objective is an end-to-end repository migration rather than implementation of a checklist-selected subset. Feature checklists are withheld during migration and are used only for post-hoc evaluation. Build feedback is available through \texttt{hvigorw assembleHap}. Build success is execution feedback and is not included in either reported metric.

All selected baseline runs use DeepSeek-V4-Pro through the one-million-token context endpoint. Prompt reuse follows each method's native interface. ReCodeAgent uses one shared four-role template with a per-application run configuration. OpenHands instantiates one repository-agent task template with source-derived application scope. RepoTransAgent uses one static contract deterministically augmented with repository structure, a grouped source inventory, selected source excerpts, a scaffold summary, and its interaction grammar. Across methods, the prompts define the same end-to-end migration objective, source-access policy, anti-placeholder requirements, and build criterion, while application-specific scope is derived from the Android source rather than from the evaluation checklist.

\subsection{Baseline Implementations}
\label{sec:baseline-implementations}

\paragraph{ReCodeAgent.} We preserve the four-role workflow of ReCodeAgent: \emph{Analyzer}, \emph{Planning}, \emph{Translator}, and \emph{Validator}. The Analyzer inspects the complete Android repository and produces source and target-design documents. The Planning agent maps source and resource groups to concrete ArkTS paths. The Translator implements the plan, and the Validator checks repository coverage, placeholders, platform residue, and build health. The Translator and Validator alternate for at most five iterations and stop early when validation passes. ReCodeAgent's original executable-test validation is not directly available for Android-to-HarmonyOS migration. Therefore, our language-pair adaptation uses the real HarmonyOS build together with static repository coverage and placeholder checks. The same four role templates are used for all repositories.

\paragraph{OpenHands.} We run OpenHands SDK 1.21.0 in headless, non-interactive mode. The general CodeAct/ReAct-style agent can inspect the source, edit the target repository, invoke shell commands, and iterate on compiler feedback. It is not assigned a fixed action budget; a run terminates when the agent emits its completion action. The prompt requires a source-derived migration inventory, a real entry flow, registered pages, non-UI subsystems, preserved application identity, and a successful HarmonyOS build. App-specific subsystem names are included only to make the source-derived scope explicit.

\paragraph{RepoTransAgent.} We instantiate the repository-level ReAct agent released with RepoTransBench. This is a single agent, not a collaborative multi-agent system. We retain its five-action interface: \texttt{ReadFile}, \texttt{CreateFile}, \texttt{ExecuteCommand}, \texttt{SearchContent}, and \texttt{Finished}. For the Android-to-HarmonyOS language pair, the prompt materializer supplies a deterministic Android repository tree, source inventory, selected source samples, and the initial HarmonyOS scaffold tree. The translation phase has a budget of 300 actions. Compiler diagnostics are then returned in up to five repair rounds of 60 actions each, for a maximum budget of 600 actions. Build success is feedback rather than an automatic stop: each phase terminates when the agent emits \texttt{Finished}, exhausts its action budget, or hits the phase timeout. In the recorded Gallery run, no phase emitted \texttt{Finished}, all 600 actions were used, and the repair phase ended with a successful build.

\section{Evaluation Details}

\subsection{Inputs and Source Grounding}

The released \texttt{a2h-evaluate} skill accepts exactly three inputs: the read-only Android repository, one read-only migrated HarmonyOS repository, and the original feature checklist. The evaluator model is recorded as execution metadata rather than treated as a fourth input. 
The selected checklist contains 323 behaviors for Gallery, 184 for AntennaPod, and 129 for Meshtastic.

The launcher first checks its local prerequisites, fingerprints the three inputs and all evaluation components, normalizes the raw checklist into stable feature rows, and creates or resumes content-addressed state. A previously generated contract is not accepted as an input. Resume is permitted only when the input and protocol fingerprints match; a mismatch is a hard failure, while an already verified prefix is preserved after interruption.

Before any HarmonyOS target file is inspected, each normalized row is grounded independently in the Android repository. The evaluator checks a valid source hint first and otherwise searches repository-relative paths, resources, control names, feature tokens, and hierarchy tokens. It traces each feature from a concrete UI, route, lifecycle callback, handler, service, wrapper, or dependency boundary through a minimal precondition and a named action or render event to the essential observable outcome.
Every row receives the smallest concrete Android \texttt{path:line} anchor and one source basis: \textsc{Exact} when app code directly supports the whole behavior, or \textsc{Boundary} when a concrete app boundary supports the feature but the checklist names a finer detail not directly encoded there. Related rows may reuse an anchor but retain distinct identities, and no selected row is removed from the evaluation scope. After all rows are drafted, the evaluator audits row identity, anchors, numeric values, and behavior boundaries; corrections are allowed only before the reference is explicitly sealed. Sealing makes the source reference immutable and records its hash, after which target judgment and alignment may begin.

\subsection{Agent-as-Judge}

For each sealed source behavior, the evaluator inspects the corresponding HarmonyOS implementation and assigns exactly one status: \textsc{Full} indicates that static code coherently implements the essential feature intent, including its principal action, observable outcome, and a statically evidenced integration path, such as a registered route, invocation, callback, lifecycle hook, exported ability, or service. Runtime proof and the exact Android UI structure are not required. \textsc{Partial} indicates that substantive code implements a meaningful part of the same intent, but integration is uncertain, or an essential action, outcome, state or data flow, or named subflow is incomplete, disconnected, mocked, or materially degraded. \textsc{Missing} indicates that no substantive implementation of the named behavior exists.

The evaluator assigns the strongest status supported by concrete static evidence. Platform-native alternatives and equivalent behavior on a different screen or navigation structure are accepted. A substantive but disconnected implementation is \textsc{Partial}, whereas names, comments, declarations, interfaces, constants, placeholders, empty callbacks, hard-coded demonstrations, mocks, boilerplate, and UI shells without the behavior are \textsc{Missing}. Static presentation counts only when the checklist row itself defines a display or information feature. One implementation may satisfy multiple rows only when it independently implements each row's essential behavior. Differences in naming, architecture, UI structure, or the absence of runtime confirmation do not by themselves cause a downgrade.

\textsc{Full} and \textsc{Partial} judgments require concrete
repository-relative \texttt{path:line} target anchors.
\textsc{Missing} requires \texttt{target\_anchor=NOT\_FOUND} and a rationale
describing the relevant target area searched. For \textsc{Partial}, the
rationale identifies both the implemented portion and the decisive remaining
gap. All checklist rows remain in the denominator. The
score is
\[  
\mathrm{Agent} =
\frac{N_{\mathrm{Full}} + 0.5N_{\mathrm{Partial}}}
{N_{\mathrm{Full}}+N_{\mathrm{Partial}}+N_{\mathrm{Missing}}}.
\]

\subsection{Node Alignment}

Node Alignment is executed only after the Android source reference has been sealed. The launcher verifies the hashes and versions of the bundled alignment tool (CodeGraph~1.4+ and its runtime) and fails rather than substituting another extractor. Each repository is copied into an indexing workspace with build and dependency directories excluded, then parsed by CodeGraph into a shared SQLite schema covering class, interface, struct, function, method, file, and enum nodes with contains, calls, instantiates, extends, implements, and references edges; instantiation is treated as a call relation, and an ArkTS class's inheritance relations are back-filled from its declaration only when CodeGraph has not emitted them.

\paragraph{Platform-agnostic abstraction.}
Both extracted graphs are independently abstracted into the same platform-agnostic semantic graph (PASG). Only architectural entities become PASG nodes, classified by deterministic platform rules into \textsc{Screen}, \textsc{DataSource}, \textsc{StateContainer}, \textsc{EventHandler}, \textsc{BackgroundTask}, \textsc{Transformer}, and \textsc{Gateway}; file- and function-level entities serve as extraction context or edge carriers. Calls and instantiations are lifted to their enclosing architectural entities, and contains, call, and inheritance relations are converted into platform-agnostic edges (\textsc{renders}, \textsc{triggers}, \textsc{observes}, \textsc{delegates}, \textsc{schedules}, \textsc{transforms}, \textsc{propagates}), with duplicates removed.

\paragraph{Platform-friction filtering.}
The Android retention denominator is counted before PASG node filtering, minus only logged platform-friction entities that require no HarmonyOS counterpart because the declarative UI and runtime subsume them (e.g., view-binding glue, \texttt{BroadcastReceiver}s, \texttt{AsyncTask}s, \texttt{RecyclerView} view holders, app-widget providers, and platform-specific image libraries). Irreducible platform differences therefore neither inflate nor depress the metric.

\paragraph{Matching.}
An automatic bootstrap first derives one-to-one cross-platform name mappings from class entities: exact-name matches, then deterministic suffix mappings (\texttt{Activity}$\rightarrow$\texttt{Page}, \texttt{Fragment}$\rightarrow$\texttt{Component}, \texttt{Service}$\rightarrow$\texttt{Ability}, \texttt{Receiver}$\rightarrow$\texttt{Handler}), then token-Jaccard matches above $0.45$. For every source--target PASG node pair, the matcher combines token Jaccard, synonym-expanded Jaccard, and character-prefix overlap into a name similarity, treating bootstrapped mappings as near-exact; cross-type pairs require a strong lexical match, and sharing the same original platform type adds a small bonus. Pairs above the frozen threshold $\theta=0.30$ become candidates, which are sorted by decreasing similarity (ties broken by identifiers) and matched by deterministic greedy one-to-one assignment, yielding a reproducible matched set $M$. A subsequent structural-refinement pass rescales retained pair scores but removes no pair and does not affect the metric.

\paragraph{Score.}
With $D_s$ the pre-filter Android retention denominator and $M$ the final matched set,
\[
\mathrm{Align} = \frac{|M|}{D_s}.
\]
The finalizer re-hashes all inputs and evaluation components, rejects any threshold change or alignment failure, and emits the two scores independently; each ``Avg.'' cell is the unweighted arithmetic mean over the three applications.

\onecolumn\section{Baseline Prompt Examples}
\label{sec:baseline-prompt-listings}

The listing below reproduces the path-sanitized Gallery prompts and reports the corresponding recorded execution flow for the three baselines. Runtime controls and resume events are labeled separately from text sent to the models. We omit only concrete paths, internal artifact filenames, complete repository and target trees, source inventories and excerpts, scaffold metadata, and long generic action examples; bracketed notes mark these omissions. The prompt bodies retain the recorded role, task, and materialized system instructions as follows.

\begin{lstlisting}[style=promptlisting,numbers=none,xleftmargin=0pt,framexleftmargin=0pt,aboveskip=3pt]
PROMPT EXAMPLE -- OpenHands

[Recorded execution flow; not part of the task prompt]
The task below was supplied once to OpenHands SDK in headless mode. The agent then inspected the source, edited the target, used shell/build feedback, and terminated with its native Finish action after a successful build. No fixed action budget was imposed.

[Redaction convention]
Concrete directory paths are replaced by semantic repository names.

TASK PROMPT

You are an expert in migrating Android applications to HarmonyOS (ArkTS / ArkUI).

Working directory layout:
- [ANDROID SOURCE] - the source Android app (Kotlin). READ-ONLY reference; do NOT modify it.
- [HARMONY TARGET] - a HarmonyOS ArkTS project scaffold. Write your translation HERE.

Goal: Translate the Android app into a functionally-equivalent HarmonyOS ArkTS app inside [HARMONY TARGET], reusing the existing scaffold.

Scope & completeness (IMPORTANT - translate the WHOLE app, not a core subset): The app's screens are the Android Activities - every Activity except the abstract base SimpleActivity. For EACH screen Activity, create a corresponding ArkTS page, register it in the page configuration, and wire navigation to it. Also translate the supporting models, adapters/components, dialogs and helpers those screens depend on. Do NOT stop at the "core" screens - the migration is complete only when EVERY screen Activity has a corresponding page.

Build & verify:
hvigorw assembleHap --no-daemon -p product=default
Fix compilation errors iteratively until the build succeeds.

Definition of success (all required):
1. Every screen Activity has a corresponding registered ArkTS page.
2. The Harmony project builds (BUILD SUCCESSFUL).
3. The app launches into its real home screen: the entry ability loads the real home page (not the placeholder), with the other screens reachable from it.

Start by listing ALL Activity source files to enumerate the screens, then translate every one, then build and fix errors.

---------------------------------------------------------------------------------------------------------------------------------------

PROMPT EXAMPLE -- RepoTransAgent

[Recorded execution flow; not part of the system prompts]
RepoTransAgent first ran one TRANSLATE phase with a 300-action limit. It then ran five REPAIR phases, each with a 60-action limit, over the same evolving target. Every repair phase used the same repair system prompt. At each turn the harness appended deterministic progress state and the latest build diagnostics. The recorded run used all 600 actions; the final repair phase ended with a successful build.

[Materialized context omitted]
The complete Android repository tree, deterministic source inventory, selected source excerpts, initial/current HarmonyOS target tree, concrete file paths, and long generic action examples are omitted. They were present in the actual materialized prompts.

TRANSLATE SYSTEM PROMPT

You are RepoTransAgent translating the entire Android repository for simple-gallery into a launchable HarmonyOS ArkTS application.

## Execution phase
translate

Source reads use source://relative/path. Relative write paths resolve inside the HarmonyOS target root. The Android source is read-only. The feature checklist and evaluator outputs are not available to you.

## End-to-end migration contract

Migrate the complete Android repository into the HarmonyOS project. The Android repository is the source of truth. Cover the launcher and app identity, UI and navigation, domain behavior, persistence, network/platform services, background work, resources, permissions, and build configuration that exist in the source.

Do not stop at a demo shell or only the visible home screen. Do not create mock data architecture, TODO/FIXME placeholders, empty handlers, or hard-coded success paths. Where Android and HarmonyOS differ, implement the closest concrete ArkTS behavior and preserve externally visible semantics.

## Translation phase

Work through the complete source inventory in coherent repository slices. Keep the launcher, domain, storage, network/platform, resource, and UI/navigation parts connected. Do not spend repeated turns re-reading the same file when a concrete source path or group from the inventory can be advanced instead.

Inspect source files as needed with ReadFile(source://...), implement directly in the target scaffold, and build frequently with ./run_tests.sh. Every page must be registered and reachable from the real entry flow. Finished(success) is valid only after a successful build following the latest target change and after the placeholder entry has been replaced by the migrated application.

## Response format

Respond with exactly one ReAct step:

Thought: [brief reasoning]
Action: [one valid action]

No text may appear outside Thought and Action.

## Available actions

CreateFile(filepath="relative/path"): Create a complete translated implementation. Implement full functionality based on source-code analysis, not an empty stub.

ReadFile(filepath="relative/path") Read a source or current target implementation needed for translation, verification, or debugging.

ExecuteCommand(command="shell command") Execute commands for build, testing, and verification.

SearchContent(keyword="search term") Search within generated target files.

Finished(status="success|failed") Mark the task complete. Use success only after verifying the latest target build.

Start by reading the launcher/entry source and its immediate dependencies, then implement the corresponding real HarmonyOS entry flow.

REPAIR SYSTEM PROMPT: PHASE-SPECIFIC PART

## Execution phase
repair

## Repair phase

The target contains the output of an earlier translation phase. Start from the current target and the latest compiler diagnostics. Repair concrete build or integration errors first. Do not restart the migration, replace working code with a demo, or declare success while the latest target change is unbuilt.

The end-to-end migration contract, source policy, build and completion requirements, materialized repository context, response format, and five-action interface are otherwise the same as in the TRANSLATE system prompt.

[Per-turn harness input]
Continue the Android-to-HarmonyOS migration in the repair phase. Inspect the existing target only as needed, fix the latest concrete build or integration problems, and keep advancing the repository instead of restarting it.

The harness then appended current iteration counts, source/target coverage, recent actions, current target files, build status, and compiler diagnostics.

---------------------------------------------------------------------------------------------------------------------------------------

PROMPT EXAMPLE -- ReCodeAgent

[Recorded execution flow; not part of a role prompt]
ReCodeAgent invoked the four roles as separate agents: Analyzer -> Planning -> Translator <-> Validator. The run allowed at most five Translator-Validator iterations and stopped when the Validator produced PASS. The recorded Gallery run completed Analyzer, Planning, and the first Translator before a phase-boundary interruption. It resumed from the first Validator without changing the role templates or wrapper, then ran three additional Translator-Validator repair cycles. The fourth Validator result was PASS.

[Redaction convention]
Concrete source, target, planning, and toolchain paths are omitted. Internal planning filenames are replaced by semantic artifact names.

ANALYZER ROLE PROMPT

# Android to HarmonyOS Repository Analysis

Analyze the complete Android repository and design its HarmonyOS/ArkTS translation.

Inspect the real repository with Read, Glob, and Grep. Cover every Kotlin source package, XML/resource family, manifest capability, dependency, and externally visible behavior. Do not use a feature checklist or ECAT judge.

You MUST write both planning artifacts:

1. [SOURCE ANALYSIS], containing the complete source tree summary, activities, adapters, dialogs, services, models, persistence, permissions, media APIs, dependencies, and resources.
2. [TARGET DESIGN], containing the target ArkTS module structure, Android-to-Harmony API mappings, translation risks, unsupported capabilities, dependency order, and build strategy.

Read enough source files to ground the documents. Finish only after both artifacts exist and contain substantive repository-level analysis.

PLANNING ROLE PROMPT

# Android to HarmonyOS Repository Translation Plan

Read [SOURCE ANALYSIS] and [TARGET DESIGN], then inspect repository files needed to validate them. Plan the entire repository, not a selected feature subset.

You MUST write [IMPLEMENTATION PLAN]. It must:

- begin with a checklist of all translation units;
- map every Android source/resource group to concrete ArkTS target paths;
- order units by dependency;
- specify Android-to-Harmony API and library mappings;
- separate source implementation, resources/UI, persistence/media, and build-validation work;
- require compilation after coherent batches;
- contain no placeholders or deferred units.

Finish only after the artifact exists and covers the complete repository.

TRANSLATOR ROLE PROMPT

# Android to HarmonyOS Repository Translation Execution

Read [IMPLEMENTATION PLAN], [SOURCE ANALYSIS], and [TARGET DESIGN]. Execute every unchecked translation unit in dependency order.

For each coherent source unit, use the source-translator-executor subagent. The subagent must inspect the corresponding Kotlin/XML source, implement real ArkTS under the target root, and preserve behavior. Do not leave TODOs, empty handlers, mocks, fake success paths, or disconnected demo screens.

After each coherent batch, and at the end, run:

hvigorw assembleHap --no-daemon -p product=default --stacktrace

Fix all compilation errors. If [VALIDATION REPORT] exists, repair every listed issue before continuing. Do not read or use the ECAT final judge, feature score, or entropy output. Finish only after all plan items are checked and the final build succeeds.

SOURCE-TRANSLATOR-EXECUTOR SUBAGENT PROMPT

Translate only the source/resource unit assigned by the parent. Inspect the real Kotlin/XML files, implement complete ArkTS in the target, and avoid TODOs, stubs, mocks, and fake success paths. Preserve existing Harmony build configuration. Run the Harmony build when the assigned unit forms a compilable batch, fix errors caused by your changes, update the implementation-plan checklist for your unit, then terminate.

VALIDATOR ROLE PROMPT

# Android to HarmonyOS Repository Validation

Read all planning documents and inspect the complete target repository. Use the validator subagents to check target structure, source-to-target coverage, missing behavior, stubs, TODOs, Android residue, identifier/API mismatches, and build health.

Run:

hvigorw assembleHap --no-daemon -p product=default --stacktrace

If any issue exists, write [VALIDATION REPORT] with status FAIL and numbered, file-specific repairs. Remove any stale validation summary. Only when the project builds and no planned unit is missing may you write [VALIDATION SUMMARY] with status PASS. Do not modify target implementation files.

STRUCTURE-VALIDATOR SUBAGENT PROMPT

Compare [IMPLEMENTATION PLAN] and the Android repository against the Harmony target. Report missing files, missing units, disconnected UI, and wrong target locations. Do not modify target files.

STUB-VALIDATOR SUBAGENT PROMPT

Inspect all target ArkTS/resources for TODO, FIXME, placeholder, mock, empty handlers, hard-coded success, unimplemented behavior, and Android API residue. Report exact files and lines. Do not modify target files.
\end{lstlisting}

\end{document}